\documentclass{article}

\usepackage{times}
\usepackage{graphicx} 
\usepackage{subfigure} 

\usepackage{amssymb}
\usepackage{amsmath}
\usepackage{color}
\usepackage{amsthm}
\usepackage{booktabs, topcapt}

\usepackage{natbib}

\usepackage{algorithm}
\usepackage{algorithmic}

\usepackage{hyperref}
\usepackage{url}

\usepackage[switch]{lineno}

\usepackage[accepted]{whi2016} 

\newcommand{\be}{\begin{eqnarray} \begin{aligned}}
\newcommand{\ee}{\end{aligned} \end{eqnarray} }
\newcommand{\benn}{\begin{eqnarray*} \begin{aligned}}
\newcommand{\eenn}{\end{aligned} \end{eqnarray*} }

\icmltitlerunning{Toward Interpretable Topic Discovery via Anchored Correlation Explanation}

\begin{document} 

\twocolumn[
\icmltitle{Toward Interpretable Topic Discovery via Anchored Correlation Explanation}

\icmlauthor{Kyle Reing}{reing@isi.edu}
\icmlauthor{David Kale}{kale@isi.edu}
\icmlauthor{Greg Ver Steeg}{gregv@isi.edu}
\icmlauthor{Aram Galstyan}{galstyan@isi.edu}
\icmladdress{USC Information Sciences Institute, Marina del Rey, CA, USA}

\icmlkeywords{interpretable machine learning, information theory}

\vskip 0.3in
]

\begin{abstract} 
Many predictive tasks, such as diagnosing a patient based on their medical chart, are ultimately defined by the decisions of human experts. 
Unfortunately, encoding experts' knowledge is often time consuming and expensive.
We propose a simple way to use fuzzy and informal knowledge from experts to guide discovery of interpretable latent topics in text. 
The underlying intuition of our approach is that latent factors should be informative about both correlations in the data and a set of relevance variables specified by an expert.
Mathematically, this approach is a combination of the information bottleneck and Total Correlation Explanation (CorEx). We give a preliminary evaluation of Anchored CorEx, showing that it produces more coherent and interpretable topics on two distinct corpora.
\end{abstract} 

\section{Introduction}
A clinician can look at a patient's electronic health record (EHR) and not only decide whether the patient has diabetes but also produce a succinct summary of the clinical evidence. Replicating this feat with computational tools
has been the focus of much research in clinical informatics.
There are major initiatives underway to codify clinical knowledge into formal representations, most often as deterministic rules that can be applied in a semi-automated fashion \cite{Newtone147}. However, representing the intuitive judgments of human experts can be challenging, particularly when the formal system does not match the expert's knowledge. For example, many deterministic disease classifiers used in clinical informatics rely heavily upon administrative codes not available at time of diagnosis.
Further, developing and testing such systems is time- and labor-intensive.

We propose instead a lightweight information theoretic framework for codifying informal human knowledge and then use it to extract interpretable latent topics from text corpora.
For example, to discover patients with diabetes in a set of clinical notes, a doctor can begin by specifying disease-specific \textit{anchor} terms \cite{arora:focs2012,halpern:amia2014}, such as ``diabetes'' or ``insulin.'' Our framework then uses these to help discover both latent topics associated with diabetes and records in which diabetes-related topics occur.
The user can then add (or remove) additional anchor terms (e.g., ``metformin'') to improve the quality of the learned (diabetes) topics.

In this workshop paper, we introduce a simple approach to anchored information theoretic topic modeling using a novel combination of Correlation Explanation (CorEx) \cite{corex} and the information bottleneck \cite{tishby}. This flexible framework enables the user to leverage domain knowledge to guide exploration of a collection of documents and to impose semantics onto latent factors learned by CorEx. We present preliminary experimental results on two text corpora (including a corpus of clinical notes), showing that anchors can be used to discover topics that are more specific and relevant. What is more, we demonstrate the potential for this framework to perform weakly supervised learning in settings where labeling documents is prohibitively expensive \cite{chen2013applying,vibhu:jamia2016}.




With respect to interpretable machine learning, our contributions are twofold.
First, our framework provides a way for human users to share domain knowledge with a statistical learning algorithm
that is both convenient for the human user and easily digestible by the machine.
Second, our experimental results confirm that the introduction of simple anchor words can improve the coherence and human interpretability of topics discovered from data.
Both are essential to successful and interactive collaboration between machine learning and human users.

\section{Methods}
\label{sec:methods}
Anchored Correlation Explanation can be understood as a combination of Total Correlation Explanation (CorEx)~\cite{corex, corex_theory} and the multivariate information bottleneck~\cite{tishby, slonimmvib}. We search for a set of probabilistic functions of the inputs  $p(y_j|x)$ for $j=1,\ldots, m$ that optimize the following information theoretic objective:
\begin{eqnarray*}
\max_{p(y_j|x), j=1,\ldots,m} TC(X;Y) + \beta \sum_{i,j \in \mathcal R} I(X_i;Y_j)
\end{eqnarray*}
The first term is the CorEx objective $TC(X;Y) \equiv TC(X) - TC(X|Y)$, which aims to construct latent variables $Y$ that best explain multivariate dependencies in the data $X$. Here the data consist of $n$-dimensional binary vectors $[X_1, \dots, X_n]$. Total correlation, or multivariate mutual information~\cite{watanabe}, is specified as $TC(X_1, \ldots, X_n) = D_{KL} \left(p(x_1,\ldots,x_n) || \prod_i p(x_i)\right)$
where $D_{KL}$ is the KL divergence. 
Maximizing $TC(X;Y)$
over latent factors $\{Y_j\}_{j=1}^m$
amounts to minimizing $TC(X|Y)$, which measures how much dependence in $X$ is explained by $Y$. At the global optimum, $TC(X|Y)$ is zero and the observations are independent conditioned on the latent factors. 
Several papers have explored CorEx for unsupervised hierarchical topic modeling~\cite{corex,chen_topic,hodas2015}.

The second term involves the mutual information between pairs of latent factors $Y_j$) and \textit{anchor} variables $X_i$ specified in the set $\mathcal R=\{(i,j)\}$.
This is inspired by the information bottleneck~\cite{tishby,slonimmvib},
a supervised information-theoretic approach to discovering latent factors. 
The bottleneck objective $\max_{p(y|x)} -I(X;Y) + \beta I(Z;Y)$ constructs latent factors $Y$ that trade off compression of $X$ against preserving information about relevance variables $Z$.

Anchored CorEx preserves information about anchors while also explaining as much multivariate dependence between observations in $X$ as possible.
This framework is flexible: we can attach multiple anchors to one factor or one anchor to multiple factors.
We have found empirically that $\beta=1$ works well and does not need to be tuned.

Anchors allow us to both seed CorEx and impose semantics on latent factors: when analyzing medical documents, for example, we can anchor a \textit{diabetes} latent factor to the word ``diabetes.'' The $TC$ objective then discovers other words associated with ``diabetes'' and includes them in this topic.

While there is not space here for a full description of the optimization,
it is similar in principle to the approaches in \citet{corex,corex_theory}. Two points are worth noting: first, the TC objective is replaced by a lower bound to make optimization feasible~\cite{corex_theory}. Second, we impose a sparse connection constraint (each word appears in only one topic) to speed up computation. Open source code implementing CorEx is available on github~\cite{sparse_code}.


%
%


\begin{figure*}[thb]
\centering
    \includegraphics[width=6.75in,clip=true,trim=0 0 0 0]{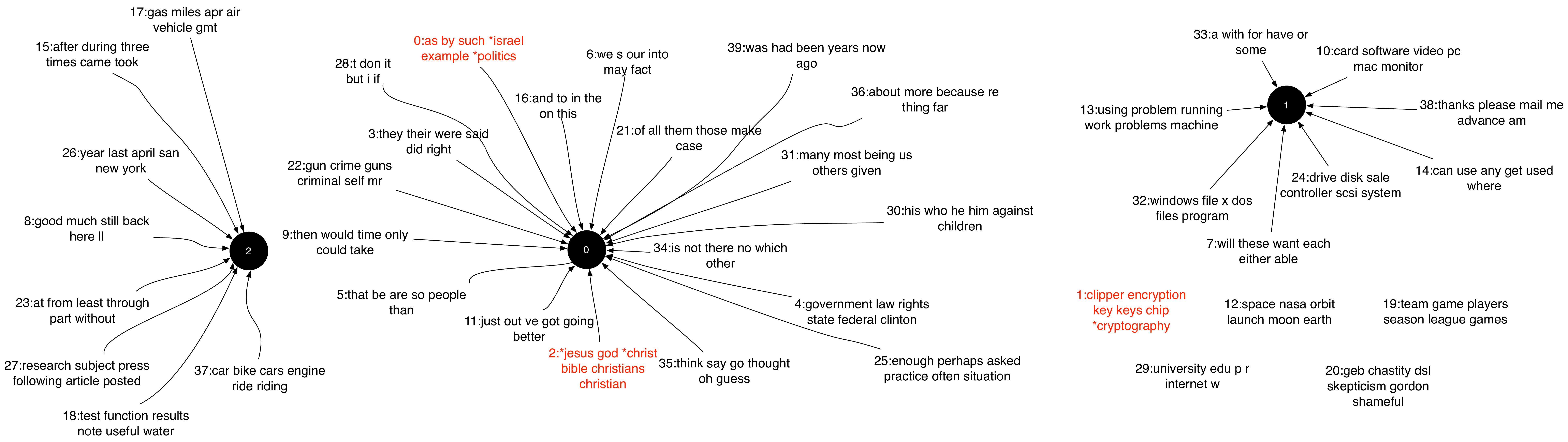} 
    \caption{A hierarchical topic model learned by CorEx. Anchored latent factors are labeled in red with anchor words marked with a ``*''.}
    \label{fig:big}
\end{figure*}

\section{Related Work}
There is a large body of work on integrating domain knowledge into topic models and other unsupervised latent variable models, often in the form of constraints \citep{wagstaff2001constrained},
prior distributions \citep{andrzejewski2009incorporating},
and token labels \citep{andrzejewski2009latent}. Like Anchored CorEx, seeded latent dirichlet allocation (SeededLDA) allows the specification of word-topic relationships \citep{jagarlamudi2012incorporating}. However, SeededLDA assumes a more complex latent structure, in which each topic is a mixture of two distributions, one unseeded and one seeded.

\citet{arora:focs2012} first proposed \textit{anchors} in the context of topic modeling: words that are high precision indicators of underlying topics. In contrast to our approach, anchors are typically selected automatically,
constrained to appear in only one topic, 
and used primarily to aid optimization \cite{nguyen2014anchors}.
In our information theoretic framework, anchors are specified manually and more loosely defined as words having high mutual information with one or more latent factors.
The effects of anchors on the interpretability of traditional topic models are often mixed \cite{lee2014low}, but our experiments suggest that our approach yields more coherent topics.

In health informatics, ``anchor'' features chosen based on domain knowledge have been used to guide statistical learning \citep{halpern:amia2014}. In \citet{vibhu:jamia2016}, anchors are used as a source of distant supervision \cite{craven:1999,mintz:2009} for classifiers in the absence of ground truth labels. While Anchored CorEx can be used for discriminative tasks, it is essentially unsupervised.
Recent work by \citet{2015arXiv151103299H} is perhaps most similar in spirit to ours: they exploit predefined anchors to help learn and impose semantics on a discrete latent factor model with a directed acyclic graph structure.
We utilize an information theoretic approach that makes no generative modeling assumptions.




\section{Results and Discussion}

To demonstrate the utility of Anchored CorEx, we run experiments on two document collections: \textit{20 Newsgroups} and the \textit{i2b2 2008 Obesity Challenge}  \cite{uzuner2009recognizing} data set. Both corpora provide ground truth labels for latent classes that may be thought of as topics.

\subsection{20 Newsgroups}
The 20 Newsgroups data set is suitable for a straightforward evaluation of anchored topic models. The latent classes represent mutually exclusive categories, and each document is known to originate from a single category. We find that the correlation structure among the latent classes is less complex than in the Obesity Challenge data. Further, each category tends to exhibit some specialized vocabulary not used extensively in other categories (thus satisfying the anchor assumption from \citet{arora:focs2012}).

To prepare the data, we removed headers, footers, and quotes and reduced the vocabulary to the most frequent 20,000 words. Each document was represented as a binary bag-of-words vector. In all experiemnts, we used the standard training/test split. All CorEx models used three layers of 40, 3, and 1 factors. \autoref{fig:big} shows an example hierarchical topic model extracted by Anchored CorEx.


\begin{table*}[thb]
\centering
\footnotesize
\begin{tabular}{lllll}
\toprule
& \multicolumn{2}{c}{\textbf{Obesity}} & \multicolumn{2}{c}{\textbf{Obstructive Sleep Apnea}} \\
\textbf{Anchors}  & \textbf{Topic} & \textbf{AUC} & \textbf{Topic} & \textbf{AUC} \\
\midrule
 & \textcolor{blue}{not fever}, \textcolor{blue}{not chill}, \textcolor{blue}{not diarrhea}, \textcolor{blue}{not dysuria}, & 0.600 & \textcolor{blue}{use}, \textcolor{blue}{drug}, \textcolor{blue}{complication}, \textcolor{blue}{allergy}, & 0.686 \\
& \textcolor{blue}{not cough}, \textcolor{blue}{not abdominal pain}, \textcolor{blue}{not guarding}, & & \textcolor{blue}{sodium}, \textcolor{blue}{infection}, furosemide, \textcolor{blue}{docusate}, & \\
& \textcolor{blue}{not rebound}, \textcolor{blue}{not palpitation}, \textcolor{blue}{not night sweats} & & shortness of breath, \textcolor{blue}{esomeprazole} & \\
&&&& \\
One per topic & \textbf{obesity}, \textcolor{red}{\textbf{sleep apnea}}, morbid obese, obese, & 0.762 & \textcolor{blue}{use}, \textcolor{blue}{complication}, \textcolor{blue}{drug}, \textcolor{blue}{allergy}, \textcolor{blue}{sodium}, & 0.546 \\
for each disease & \textcolor{blue}{labor}, acebutolol, \textcolor{blue}{vaginal bleeding},&& \textcolor{blue}{infection}, furosemide, \textcolor{blue}{docusate}, & \\
&  klonopin, valproic acid, \textcolor{blue}{bacteruria} && shortness of breath, \textbf{obstructive sleep apnea} & \\
&&&& \\
Add second OSA & \textbf{obesity}, morbid obese, obese, \textcolor{blue}{labor}, & 0.757 &  \textbf{sleep apnea}, \textbf{obstructive sleep apnea}, & 0.826 \\
anchor & not non-compliant, acebutolol, && oxygen, duoneb, desaturation, singulair, & \\
& \textcolor{blue}{vaginal bleeding}, \textcolor{blue}{problem}, & & pulmonary hypertension, hypoxemia, &\\
& not deep venous thrombosis, overweight && \textcolor{blue}{pap smear}, vicodin & \\
\bottomrule
\end{tabular}
\caption{Evolution of Obesity and Obstructive Sleep Apnea (OSA) topics as anchors are added. Colors and font weight indicate \textbf{anchors}, \textcolor{blue}{spurious terms}, and \textcolor{red}{intruder terms from other known topics}. Multiword and negated terms are the result of the preprocessing pipeline.}
\label{tab:obesity:topics}
\end{table*}

\subsection{i2b2 Obesity Challenge 2008}
The Obesity Challenge 2008 data set\footnote{ \url{https://www.i2b2.org/NLP/DataSets/Main.php}}
includes 1237 deidentified clinical discharge summaries from the Partners HealthCare Research Patient Data Repository. All summaries have been labeled by clinical experts with obesity and 15 other conditions commonly comorbid with obesity, ranging from Coronary Artery Disease (663 positives) to Depression (247) to Hypertriglyceridemia (62).

We preprocessed each document with a standard biomedical text pipeline that extracts common medical terms and phrases (grouping neighboring words where appropriate) and detecting negation (``not'' is prepended to negated terms) \citep{dai2008efficient,CHAPMAN2001301}. We converted each document to a binary bag-of-words with a vocabulary of 4114 (possibly negated) medical phrases. We used the 60/40 training/test split from the competition.

We are primarily interested in the ability of Anchored CorEx to extract latent topics that are unambiguously associated with the 16 known conditions. We train a series of CorEx models with 32 latent topics in the first layer, each using a different anchor strategy. \autoref{tab:obesity:topics} shows the Obesity and Obstructive Sleep Apnea (OSA) topics for three iterations of Anchored CorEx with the ten most important terms (highest weighted connections to the latent factor) listed for each topic. Unsupervised CorEx (first row) does not discover any topics obviously related to obesity or OSA, so we choose the topics to which the terms \textit{obesity} and \textit{obstructive sleep apnea} are assigned.
No unambiguous Obesity or OSA topics emerge even as the number of latent factors is decreased or increased.

In the second iteration (second row), we add the common name of each of the 16 diseases as an anchor to one factor (16 total). Adding \textit{obesity} as an anchor produces a clear Obesity topic, including several medications known to cause weight gain (e.g., \textit{acebutolol}, \textit{klonopin}). The anchored OSA topic, however, is quite poor and in fact resembles the rather generic topic to which \textit{obstructive sleep apnea} is assigned by Unsupervised CorEx. It includes many spurious or non-specific terms like \textit{drug}.

This is likely due to the fact that obesity is a major risk factor of OSA, and so OSA symptoms are highly correlated with obesity and its other symptoms. Thus, the total correlation objective will attempt to group obesity and OSA-related terms together under a single latent factor. The sparse connection constraint mentioned in \autoref{sec:methods} prevents them from being connected to multiple factors.
Indeed, \textit{sleep apnea} appears in the obesity topic, suggesting the two topics are competing to explain OSA terms. 

In the third iteration, we correct this by adding \textit{sleep apnea} as a second anchor to the OSA topic, and the resulting topic is clearly associated with OSA, including terms related to respiratory problems and medications used to treat (or believed to increase risk for) OSA. There is no noticeable reduction in quality in the Obesity topic.   

\subsection{Anchored CorEx for Discriminative Tasks}
In a series of follow-up experiments, we investigate the suitability of using anchored CorEx to perform weakly supervised classification. We interpret each anchored latent factor as a classifier for an associated class label and then compute test set F1 (using a threshold of 0.5) and area under the curve (AUC) scores (Obesity Challenge only).

\begin{table}[htbp]
   \centering
   \begin{tabular}{@{} lccc @{}} 
   
      \toprule
\textbf{Anchors}&\textbf{$F1_{Anch}$}&\textbf{$F1_{Unsup}$}\\ 
      \midrule

Jesus&0.42&0.45\\
God&0.49&0.43\\
Jesus,Christian&\textbf{0.55}&0.45\\
\midrule
Naive Bayes&\multicolumn{2}{c}{0.75} \\
      \bottomrule
   \end{tabular}
\caption{F1 scores on \texttt{soc.religion.christianity}.}
   \label{tab:class}
\end{table}

\autoref{tab:class} compares the classification performance of Unsupervised and Anchored CorEx on the \texttt{soc.religion.christianity} category from 20 Newsgroups for different choices of anchors. For both types of CorEx, the topic containing the corresponding terms is used as the classifier, but for Anchored CorEx those terms are also used as anchors when estimating the latent factor. Unsupervised CorEx does a reasonable job of discovering a coherent religion topic that already contains the terms \textit{God}, \textit{Christian}, and \textit{Jesus}. However, using the terms \textit{Jesus} and \textit{Christian} as anchors yields a topic that better predicts the actual \texttt{soc.religion.christianity} category.\\

\begin{table}[ht]
\centering
\begin{tabular}{lll}
\toprule
\textbf{Classifier} & \textbf{Macro-AUC} & \textbf{Macro-F1} \\
\midrule
Naive Bayes & 0.7120 & 0.4638 \\
Anchored CorEx & 0.7445 & 0.5328 \\
\bottomrule
\end{tabular}
\caption{Classification performance on Obesity 2008.}
\label{tab:obesity:class}
\end{table}
\vspace{-10pt}

\autoref{tab:obesity:class} shows the Macro-AUC and F1 scores (averaged across all diseases) on the Obesity Challenge data for the final anchored CorEx model and a Naive Bayes (NB) baseline, in which we train a separate classifier for each disease.
Surprisingly, Anchored CorEx outperforms Naive Bayes (NB) by a large margin.
Of course, Anchored CorEx is not a replacement for supervised learning: NB beats Anchored CorEx on 20 Newsgroups and does not represent a ``strong'' baseline for Obesity 2008 (teams scored above 0.7 in Macro-F1 during the competition). It is nonetheless remarkable that Anchored CorEx performs as well as it does given that it is fundamentally unsupervised.


\section{Conclusion}
We have introduced a simple information theoretic approach to topic modeling that can leverage domain knowledge specified informally as \textit{anchors}. Our framework uses a novel combination of CorEx and the information bottleneck. Preliminary results suggest it can extract more precise, interpretable topics through a lightweight interactive process. We next plan to perform further empirical evaluations and to extend the algorithm to handle complex latent structures present in health care data.

\subsection*{Acknowledgements}
This work was partially supported by DARPA award HR0011-15-C-0115. David Kale was supported by the Alfred E. Mann Innovation in Engineering Doctoral Fellowship.

\clearpage
\small
\bibliography{gversteeg_bibdesk,more_refs}
\bibliographystyle{icml2016}


\end{document}